\documentclass[letterpaper, 10 pt, conference]{ieeeconf} 
\IEEEoverridecommandlockouts  

\usepackage{graphicx} % Required for inserting images

\usepackage{amsmath,amssymb,amsfonts,soul}

\usepackage{graphicx}
\usepackage{textcomp}
\usepackage[normalem]{ulem}

\usepackage{amsthm}
\usepackage{graphics,wrapfig} % for pdf, bitmapped graphics files
\usepackage{epsfig} % for postscript graphics files
\usepackage{times} % assumes new font selection scheme installed

\usepackage[maxbibnames=6,giveninits=true]{biblatex}
\usepackage{bm}
\usepackage{xcolor}
\usepackage{booktabs}
\usepackage{siunitx}
\usepackage{hyperref}

\usepackage{caption}
\usepackage{lipsum}% Example text
\usepackage{multirow}
\usepackage{diagbox} 
\usepackage{subcaption}
\usepackage{tabularx}
\usepackage{slashbox} % diagonal slash in table

\newcommand{\R}{\mathbb{R}}%commands for easy math notations

\newcommand{\N}{\mathcal{N}}
\newcommand{\Pro}{\mathbb{P}}
\newcommand{\io}{\iota_{\varepsilon}}

\newcommand{\sigi}{\sigma_i}

\newcommand{\dd}{\textup{d}}
\newcommand{\bp}{\textup{ \bf p}}

\usepackage[T1]{fontenc}
\newcommand{\textincon}[1]{%
{\fontfamily{zi4}\selectfont #1}}
\newcommand{\VAR}{{\textrm{\large \textincon{V$@$R}}}}
\newcommand{\AVAR}{{\textrm{\large \textincon{AV$@$R}}}}
\def\bp{{\bf p}}

\newtheorem{definition}{\bf Definition}

\newtheorem{theorem}{\bf Theorem}
\newtheorem{lemma}{\bf Lemma}
\newtheorem{corollary}{\bf Corollary}

\def\bx{{\bf x}}
\def\by{{\bf \Phi}}
\def\bxa{{\bf x}^{acq}_i}
\def\bya{{\bf \Phi}^{acq}_i}
\def\bw{{\bf w}}
\def\bc{{\bf c}}
\def\bH{{\mathbf H}}
\DeclareMathOperator{\tr}{tr}

\DeclareMathOperator*{\argmax}{arg\,max}
\DeclareMathOperator*{\argmin}{arg\,min}

\DeclareMathOperator{\diag}{diag}

\title{\bf Beyond Uncertainty: Risk-Aware Active View Acquisition for Safe Robot Navigation and 3D Scene Understanding with FisherRF}

\author{Guangyi Liu$^*$, Wen Jiang$^*$, Boshu Lei$^*$, Vivek Pandey, Kostas Daniilidis, and Nader Motee% <-this % 
\thanks{$*$ G. Liu, W. Jiang, and B. Lei contributed equally to this work.\endgraf
G. Liu, V. Pandey, and N. Motee are with the Department of Mechanical Engineering and Mechanics, Lehigh University, Bethlehem, PA 18015, USA. {\tt\small \{gliu,vkp219,motee\}@lehigh.edu}.\endgraf
W. Jiang, B. Lei, and K. Daniilidis are with the Department of Computer and Information Science, University of Pennsylvaniac, Philadelphia, PA 19104, USA. {\tt\small \{wenjiang,leiboshu\}@seas.upenn.edu}, {\tt\small \{kostas\}@cis.upenn.edu}
}
}

\usepackage{geometry}
\begin{document}

\maketitle

\thispagestyle{plain}
\pagestyle{plain}
% \keywords{LaTeX, document formatting, keywords, research paper}

\begin{abstract}
The active view acquisition problem has been extensively studied in the context of robot navigation using NeRF and 3D Gaussian Splatting. To enhance scene reconstruction efficiency and ensure robot safety, we propose the Risk-aware Environment Masking (RaEM) framework. RaEM leverages coherent risk measures to dynamically prioritize safety-critical regions of the unknown environment, guiding active view acquisition algorithms toward identifying the next-best-view (NBV). Integrated with FisherRF, which selects the NBV by maximizing expected information gain, our framework achieves a dual objective: improving robot safety and increasing efficiency in risk-aware 3D scene reconstruction and understanding. Extensive high-fidelity experiments validate the effectiveness of our approach, demonstrating its ability to establish a robust and safety-focused framework for active robot exploration and 3D scene understanding.
\end{abstract}

%%%%%%%%%%%%%%%%%%%%%%%%%%%%%%%%%%%%%%%%%%%%%%%%%%%%%%%%%%%%%%%%%%%%%%%%%%%%%%%%%%%%%
\section{Introduction}

The growing need for robots to function effectively in unpredictable and dynamically changing environments, such as disaster zones, and search $\&$ rescue operations, necessitates significant advancements in active perception. In these dynamic environments, robots face a dual challenge: exploring their surroundings to accurately reconstruct the environment, e.g., Neural Radiance Field (NeRF) models \cite{mildenhall2020nerf} and 3D Gaussian Splatting \cite{ keetha2024splatam}, while prioritizing their safety and ensuring mission success \cite{georgakis2022uncertainty}. To obtain a higher-quality reconstruction, robots actively gather additional samples from the environment, known as active view acquisition for 3D reconstruction, scene understanding, and robotic vision. Existing methods for active view acquisition \cite{pan2022activenerf,ran2023neurar,zhan2022activermap,yan2023active} often prioritize the overall efficiency of scene reconstruction. However, these approaches often neglect crucial safety considerations, potentially exposing the robot to environmental hazards or compromising its operational integrity.

This work, we use the 3D Gaussian Splatting  \cite{keetha2024splatam} for initial 3D scene construction. We then incorporate the perceived environmental map while simultaneously evaluating its inherent uncertainty, similar to the approach presented in \cite{jiang2024fisherrf}. By applying notions of risk measures \cite{rockafellar2002conditional,liu2023risk} to this uncertain environment, we can effectively quantify the risk associated with different areas. Our approach leverages a technique called Risk-aware Environment Masking (RaEM) to prioritize and focus on potential viewpoints in high-risk areas of the environment. This is achieved by effectively masking out areas deemed safe. We then utilize an approach similar to FisherRF \cite{jiang2024fisherrf} to select the next-best-view (NBV) from a set of candidate viewpoints. This selection process maximizes the expected information gain, focusing on areas with the most significant potential for reducing uncertainty. Through these combined steps, we aim to refine the existing methods for the NBV selection. Our method dynamically adjusts the focus of the perception model based on the anticipated risk the robot faces in an uncertain environment.

In our proposed methodology, we convert the perceived 3D environment, represented by 3D Gaussian point clouds, into random variables, capturing the distance between the robot and different parts. This allows us to quantify the average value-at-risk, denoted by $\AVAR$, of collision for each region. By integrating RaEM with extensive simulations, the calculated $\AVAR$ values achieve a closer resemblance, quantified by the Wasserstein distance, to the ground truth risk compared to existing methods. This signifies that RaEM offers a more accurate assessment of the risk of potential collisions with the environment. 

A summary of our contributions includes a novel approach that significantly enhances existing active view acquisition methods for robots. This is achieved by integrating risk assessment directly into the exploration process. Our approach goes beyond merely understanding the environment's uncertainty such that it also evaluates potential risks associated with different waypoints, allowing robots to operate more safely. Through the combined application of RaEM and FisherRF, we equip robots with a more comprehensive awareness of their surroundings. This enables them to navigate safely and make informed decisions in real-world scenarios. All the proofs of the theoretical results are provided in the appendix.

%%%%%%%%%%%%%%%%%%%%%%%%%%%%%%%%%%%%%%%%%%%%%%%%%%%%%%%%%%%%%%%%%%%%%%%%%%%%%%%%%%%%%
\section{Mathematical Notations} 

\begin{figure*}
    \centering
    \includegraphics[width=\linewidth]{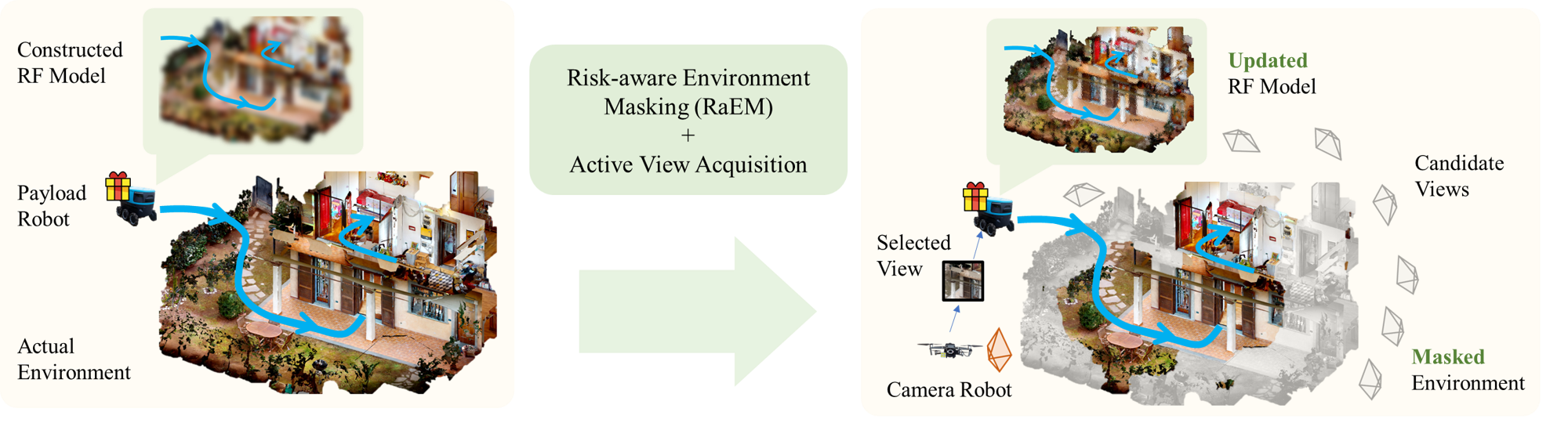}
    \caption{The payload robot aiming to deliver the asset while the camera robot aims to find the NBV to improve the risk assessment.}
    \label{fig:diag}
    \vspace{-0.4cm}
\end{figure*}

Throughout the paper, $\mathbb{R}^n$ denotes the $n$-dimensional Euclidean space and $\mathbb{R}^n_{+}$ its non-negative orthant. The identity matrix is represented as $I_n \in \mathbb{R}^{n \times n}$. We use $\|x\|_2$ for the Euclidean $2$-norm of a vector $x$. The cardinality of a set $\mathcal X$ is denoted by $|\mathcal X|$. The set $\mathcal{L}^{2}(\mathbb{R}^{n})$ consists of all $\mathbb{R}^n$-valued random vectors in a probability space $(\Omega, \mathcal{F}, \mathbb{P})$ with finite second moments. A normal random variable $\bm{y} \in \mathbb{R}^{n}$, with mean $\bm{\mu} \in \mathbb{R}^{n}$ and covariance matrix $\Sigma \in \mathbb{R}^{n \times n}$, is denoted as $\bm{y} \sim \mathcal{N}(\bm{\mu}, \Sigma)$. The error function $\text{erf}:\mathbb{R} \rightarrow (-1,1)$ is defined by
$\text{erf}(x) = \frac{2}{\sqrt{\pi}} \int_{0}^{x} e^{-t^2} \text{d} t,$ and is invertible as $\text{erf}^{-1}(x)$ within its range.

%%%%%%%%%%%%%%%%%%%%%%%%%%%%%%%%%%%%%%%%%%%%%%%%%%%%%%%%%%%%%%%%%%%%%%%%%%%%%%%%%%%%%
\section{Problem Formulation}

We consider a primary payload robot tasked with transporting and delivering an asset to a designated target location. The robot navigates along a predefined path, $\bp = (p_1, p_2, \ldots, p_N)$, where each waypoint $p_i$ lies in $\mathbb{R}^3$. The robot's onboard processor generates actions to move sequentially toward the next waypoint. The map of the unknown environment is constructed and stored on the payload robot, which has sufficient computational resources to perform onboard training for 3D Gaussian Splat \cite{keetha2024splatam}. The training is completed with a set of sampled views from various viewpoints within the environment.

Given the limited scope of the training data (sampled views), the current model inherently carries a degree of uncertainty. This uncertainty can introduce bias into the safety and risk assessment of the predefined path due to inaccuracies in the resulting map. Consequently, even if the path satisfies the safety criteria based on the current environmental map, there remains a non-negligible risk that the payload robot could still collide with obstacles and fail to deliver the asset.

Therefore, obtaining a more accurate risk assessment is crucial, as it enables the payload robot to choose a more efficient path with a reduced likelihood of violating safety constraints. It also prevents underestimating potential dangers, which could lead to failure in actual deployment. To improve safety and mitigate biased risk assessments, an auxiliary camera robot, potentially an aerial one, is deployed to the environment to gather additional views. These views are then shared with the payload robot to update its onboard 3D Gaussian Splat. This process is illustrated in Fig. \ref{fig:diag}. Given the time constraints and limited onboard computational power, the camera robot selects the NBV from a set of candidate poses, rather than exploring all possible viewpoints.

Improving the safety and risk assessment of the payload robot's navigation can be achieved by incorporating additional environmental views during the training of the 3D Gaussian splatting model. However, due to limited sensing resources (e.g., time and energy), the primary \textit{challenge} is to efficiently identify the risk-aware next best view. This approach focuses on refining the environmental map by prioritizing areas most critical to enhancing the robot's safety measures.

%%%%%%%%%%%%%%%%%%%%%%%%%%%%%%%%%%%%%%%%%%%%%%%%%%%%%%%%%%%%%%%%%%%%%%%%%%%%%%%%%%%%%
\section{Technical Background}

In this section, we provide an overview of the scene reconstruction model and delve into the risk measures for safety evaluation.

\subsection{Volumetric Rendering and Gaussian Splatting}
3D scene reconstruction techniques like 3D Gaussian Splatting \cite{kerbl3Dgaussians}, Plenoxels \cite{plenoxels} leverage volumetric rendering \cite{kajiya1984ray} for effectively learning a 3D scene representation from photo-metric supervision.  
% Consider a camera ray $\text{r}(t) = \text{o} + t\text{d}$ originating from the camera center $\text{o}\in \mathcal{R}^3$ and passing through a specific pixel on the image plane. 
The pixel color $C(\text{r})$ is determined by integrating colors $\text{c}(\text{r}(t))$ weighted by the density $\rho(\text{r}(t))$ along the path of a camera ray $\text{r}(t) = \text{o} + t \, \text{d}$, which emits from the camera's origin $\text{o} \in \mathbb{R}^3$ and intersects a particular pixel on the image plane, as described by:
\begin{equation}
    \setlength{\abovedisplayskip}{1ex}
    C(\text{r}) = \int_{t_n}^{t_f} T(t) \, \rho(\text{r}(t)) \, \text{c}(\text{r}(t),\text{d})\, dt\, , 
    \setlength{\belowdisplayskip}{1ex}
\end{equation}
where $T(t)=\exp\left(-\int_{t_n}^t\rho(\text{r}(s))\, ds\right)$ represents the cumulative transmittance with $t_n$ and $t_f$ denoting the scene's near and far limits, respectively. NeRF \cite{mildenhall2020nerf} approximates this integral via stratified sampling, expressing it as a weighted sum of sampled points:
\begin{align}    \label{eq:render}
    \hat{C}(\text{r}) &= \sum_{i=1}^{N_s} T_i \big(1-\exp(-\rho_i \, \delta_i) \big) \bc_i,\\
    T_i &=\text{exp}\bigg(-\sum_{j=1}^{i-1}\rho_j\delta_j\bigg),    
\end{align}
where $\delta_i=t_{i+1}-t_i$ calculates the interval between consecutive samples, with $N_s$ specifying the sample count. Through this methodology, Radiance Field models refine the 3D scene representation by minimizing the reconstruction errors between the actual captured RGB images ${\mathcal{I}_{i=1}^N}$ and the computed pixel colors. Although the rendering approach of 3D Gaussian Splatting is distinct from previous NeRF methods, its image formation equation still follows this structure, wherein $\textbf{c}_i$ denotes the color of each 3D Gaussian, and $\rho_i$ is determined by a 2D Gaussian with covariance $\Sigma$ \cite{kerbl3Dgaussians,yifan2019differentiable}.
Following \cite{keetha2024splatam}, we initialize new 3D Gaussians by unprojecting the RGB-D data on regions outside the rendered silhouette. We also simplified the 3D Gaussian to be isotropic such that its opacity is given by:
\begin{equation}
    \rho_i = \epsilon_i \exp \left(- \frac{||x_i - \mu||^2}{2 r^2}\right),
\end{equation}
where $x_i \in \mathbb{R}^2$ is the location of the pixel, $\epsilon$ is the opacity of the 3D Gaussian and $r$ is the scale of 2D Gaussian projected to the screen. Our scene reconstruction system takes both RGB and depth as input to densely supervise 3D Gaussians. The depth of a pixel is rendered as the weighted sum of depths $d_{dep,i}$ of 3D Gaussians:
% \wen{TODO(wen): provide details of 3D Gaussian and SplaTAM  as well}
\begin{equation}
\label{eq:render-depth}
\mathbf{d}_{dep}(\text{r}) = \sum_{n=1}^{N_s} T_i \left(1-\exp(-\rho_n \delta_n) \right) d_{dep,i}).
\end{equation}

The scene is optimized with both the RGB rendering loss and the depth supervision:
\begin{equation}
    \mathcal{L} = \sum_{r\in R} \big(\mathcal{L}_1( C(\text{r})) + \gamma  \mathcal{L}_1(\mathbf{d}_{dep}(\text{r})) \big)
    \label{eq:loss}
\end{equation}
where $\mathcal{L}_1(C(r))$ and $ \mathcal{L}_1(\mathbf{d}_{dep}(\text{r}))$ are $L1$ loss for RGB and depth respectively, and $\gamma$ is the weight for the depth loss.

\subsection{Measuring Risk of Collision} 

To quantify the uncertainty and safety associated with the relative distances between a specific waypoint (the payload robot’s position) and mass points in the environment, modeled as 3D Gaussians, we employ the widely recognized Average Value-at-Risk ($\AVAR$), also known as Conditional Value-at-Risk \cite{rockafellar2002conditional}. $\AVAR$ provides an estimate of the proximity between the robot and obstacles in the most critical scenarios, offering valuable insights into the severity of potential near-collision events. For example, with a confidence level of $\varepsilon = 5\%$, $\AVAR$ calculates the mean relative distance in the direst $5\%$ of cases, representing the scenarios where a collision is most likely to occur.

Within the probability space $(\Omega, \mathcal{F}, \mathbb{P})$, $\AVAR$ \cite{sarykalin2008value} of the continuous random variable $y: \Omega \rightarrow \mathbb{R}$, is defined as
\begin{equation}    \label{eq:avar_def}
    \AVAR_\varepsilon(y) := \mathbb{E} \left[ y \, \big| \, y < \VAR_\varepsilon(y) \right],
\end{equation}
where the value-at-risk ($\VAR$) is specified as\footnote{Contrary to the conventional definition \cite{rockafellar2002conditional} $\scalebox{0.8}{$\VAR$}_\varepsilon := \inf \left\{ z \, | \, \mathbb{P} \left\{ y > z  \right\} < \varepsilon \right\}$, it is necessary to consider the opposite tail of the distribution when contemplating collision events.}:
\begin{equation}   \label{eq:var_def}
    \VAR_\varepsilon(y) := \inf \left\{ z \, \big| \, \mathbb{P} \left\{ y < z  \right\} > \varepsilon \right\},
\end{equation}
with $\varepsilon \in (0,1)$ denoting the confidence level. When the random variable $y$ characterizes the distance, a smaller value of $\AVAR_\varepsilon$ indicates a higher chance the collision might happen.

%%%%%%%%%%%%%%%%%%%%%%%%%%%%%%%%%%%%%%%%%%%%%%%%%%%%%%%%%%%%%%%%%%%%%%%%%%%%%%%%%%%%%
\section{Collision Risk Quantification}

The risk of collision between the payload robot and the uncertain environment can be quantified by evaluating the uncertain distance between each waypoint and the perceived 3D Gaussians, as detailed below.

\subsection{Statistics of the Distance to 3D Gaussians}
\begin{figure}
    \centering
    \includegraphics[width=\linewidth]{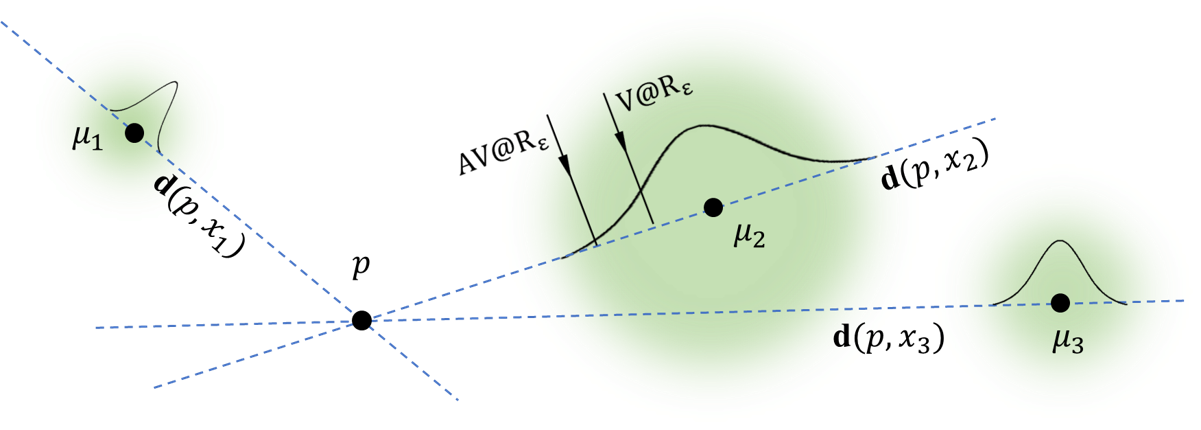}
    \caption{Distance from waypoint $p$ to 3D Gaussians $x_i \in \mathcal{X}$.}
    \label{fig:distance}
\end{figure}

We represent the 3D Gaussian model of the environment as a discrete set $\mathcal{X}$, consisting of independent multivariate normal random variables $x_i \in \mathbb{R}^3$, for $i = 1, \ldots, |\mathcal{X}|$, where $|\mathcal{X}|$ denotes the total number of perceived Gaussians. As outlined in \cite{keetha2024splatam}, each point $x_i$ is characterized by $x_i \sim \mathcal{N}(\mu_i, \Sigma_i)$\footnote{This notation focuses solely on the position and radius-related parameters of the 3D Gaussians, excluding the color-related attributes discussed in \cite{keetha2024splatam}.}, where $\mu_i \in \mathbb{R}^3$ is the mean (center) position, and $\Sigma_i = \sigma_i^2 I_3$ is the covariance matrix. These parameters are stored in the payload robot's onboard memory. We then define a random variable to quantify the distance between a given waypoint and each 3D Gaussian point (Fig.~\ref{fig:distance}).

\begin{definition}      \label{def:distance}
The signed distance between a fixed waypoint $p \in \R^3$ and any 3D Gaussian $x_i$ is given by
\[
    {\textup{d}}(p,x_i) = \left\langle x_i - p \, , \frac{\mu_i - p}{\|\mu_i - p\|_2} \right\rangle \in \R,
\]
where $\langle . , . \rangle$ is the Euclidean inner product. 
\end{definition}

Furthermore, we demonstrate that \(\textup{d}(p, x_i)\) is a normal random variable, as formalized in the following lemma.

\begin{lemma}  \label{lem:prob_dist_d(x,p)}  
Let \(p \in \mathbb{R}^3\) be a fixed waypoint, and let \(x_i \sim \mathcal{N}(\mu_i, \sigma_i^2 I_3)\) be a 3D Gaussian with $x_i \in \mathcal{X}$. The signed distance \(\textup{d}(p, x_i)\) follows a normal distribution:  
\[
    \textup{d}(p, x_i) \sim \mathcal{N} \left( \|\mu_i - p\|_2, \, \sigma_i^2 \right).
\]  
\end{lemma}  

According to Definition \ref{def:distance}, the signed distance $\textup{d}(p, x_i)$ can take negative values. This implies that a realization of the 3D Gaussian $x_i$ may lie on the side of $p$ opposite to the mean $\mu_i$ in the space. A more general result, covering the case of an anisotropic covariance matrix, is provided in Appendix.

\subsection{Risk of Collision with the Uncertain Environment}

Since $\dd(p, x_i)$ is a continuous random variable, we can quantify the average value-at-risk of a collision with a single 3D Gaussian point as follows.

\begin{theorem} \label{thm:avar}
Let $p \in \mathbb{R}^3$ be a waypoint, and let $x_i \sim \mathcal{N}(\mu_i, \sigma_i^2 I_3)$ represent a 3D Gaussian point. The average value-at-risk ($\AVAR$) of a collision at waypoint $p$ with respect to the 3D Gaussian $x_i$ is given by:
\begin{equation} \label{eq:avar}
    \AVAR_\varepsilon \left( \textup{d}(p, x_i) \right) := \|\mu_i - p\|_2 - \frac{\sigma_i}{\sqrt{2\pi}} \frac{1}{\varepsilon \exp(\iota^2)},
\end{equation}
where $\iota = \textup{erf}^{-1}(2\varepsilon - 1)$ and $\varepsilon \in (0, 1)$ is the desired confidence level.
\end{theorem}

The value of $\AVAR_\varepsilon(\dd(p, x_i))$ represents the expected conditional distance when the signed distance between the waypoint and the 3D Gaussian point falls below a specified threshold, $\VAR_\varepsilon$, as shown in Fig.~\ref{fig:distance}. A lower $\AVAR_\varepsilon(\dd(p, x_i))$ indicates a higher risk of collision between the payload robot and the 3D Gaussian point.

For any given waypoint $p_k$ in $\mathbf{p}$, there are $|\mathcal{X}|$ associated risk values, each corresponding to a distinct 3D Gaussian point $x_i$. In this paper, we focus on the most hazardous scenario, characterized by the lowest $\AVAR_\varepsilon$, to quantify the risk of collision with the uncertain environment at $p_k$. This is defined as:
\begin{equation} \label{eq:closest_dist} 
\alpha_k = \min_{x_i \in \mathcal{X}} ~\AVAR_\varepsilon \big(\dd(p_k, x_i)\big), \end{equation}
which represents the worst-case collision risk within the current perceived environment.

%%%%%%%%%%%%%%%%%%%%%%%%%%%%%%%%%%%%%%%%%%%%%%%%%%%%%%%%%%%%%%%%%%%%%%%%%%%%%%%%%%%%%
\section{Risk-aware Environment Masking and the Risk-aware Next Best View}

Finding the next best view with respect to the entire environment \cite{jiang2024fisherrf} can implicitly improve risk assessment during the payload robot's navigation but often leads to resource waste on unrelated regions. In contrast, our approach targets a specific set of 3D Gaussian points near each waypoint by masking irrelevant regions and selecting the next best view within a reduced, focused subset of the environment, thereby enhancing both efficiency and effectiveness.

\subsection{Risk-Aware Environment Masking} 
\begin{figure}
    \centering
    \includegraphics[width=\linewidth]{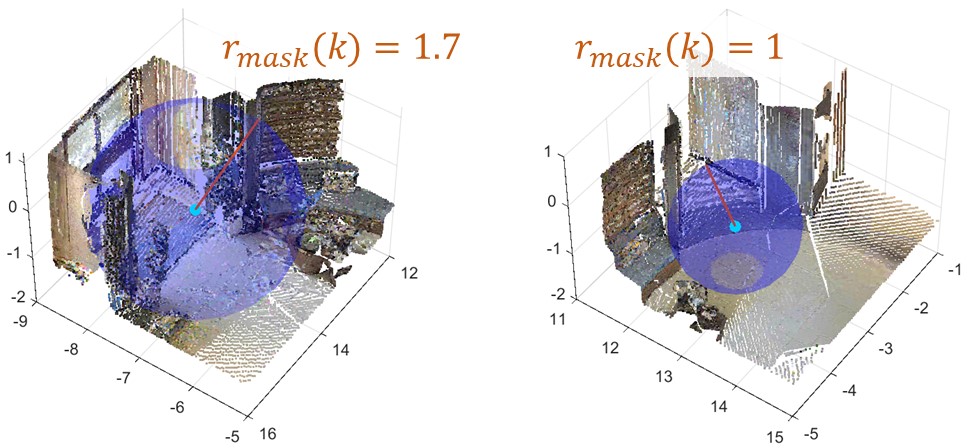}
    \caption{Masking radius of RaEM.}
    \label{fig:masking_radius}
\end{figure}

Risk-aware Environment Masking (RaEM) operates by defining a sphere around each waypoint $p_k$ along the path. This sphere, with a radius $r_{\text{mask}}(k) > 0$, is tailored to the potential risk posed by the uncertain environment at that waypoint. The radius reflects the level of risk to the robot, ensuring that only the most relevant areas are considered.

\begin{definition}[Masking Radius]
    The masking radius at each waypoint $p_k$ is defined as:
    \begin{equation} \label{eq:radius}
        r_{\text{mask}}(k) = \beta_1 \cdot e^{-\beta_2 \cdot \alpha_k},
    \end{equation}
    where $\beta_1$ and $\beta_2$ are positive constants that determine the sensitivity of the masking radius to the risk level $\alpha_k$. Examples of these parameters are provided in Sec.~\ref{sec:experiments}.
\end{definition}

A key feature of RaEM is its dynamic masking radius, which adjusts according to the robot’s perceived risk at each waypoint. In high-risk situations (indicated by a lower $\alpha_k$ value), the radius expands, prompting the robot to focus cautiously on a larger area around the waypoint for a more thorough assessment. Conversely, in lower-risk scenarios (higher $\alpha_k$ value), the radius contracts, allowing the robot to explore a broader area while minimizing attention to less critical regions. This adaptive approach ensures that safety-critical zones in the environment are prioritized, as illustrated in Fig.~\ref{fig:masking_radius}. Mathematically, the masked environment produced by RaEM is a subset of the original environment, containing only the Gaussians that lie within the union of spheres centered at each waypoint. 

For a given path ${\bf p}=(p_1, \ldots, p_N)$, these spheres have radii determined by the masking function, and the preserved 3D Gaussians (environment) are given by:
\begin{equation} \label{eq:masking}
    \mathcal{M}({\bf p}) := \left\{x \in {\mathcal{X}} ~\Bigg|~ x \in \bigcup_{k=1}^{N} \mathcal{B} \Big(p_k,r_{\text{mask}} (k) \Big) \right\} \subseteq \mathcal{X},
\end{equation}
where $\mathcal{B}(p_k, r_{\text{mask}} (k))$ represents a Euclidean ball centered at $p_k$ with radius $r_{\text{mask}} (k)$. This dynamic masking process refines the robot's focus based on the perceived risk level, enabling a more efficient and risk-aware navigation strategy.

\subsection{Next Best View Selection with RaEM}

Having characterized the masked environment, our next step involves quantifying the informativeness of each candidate view. The aim is to select the most informative view that will improve the accuracy of risk assessment for the close proximity to the path $\bf{p}$. We utilize the Fisher information to quantify the Expected Information Gain (EIG) at each candidate view~\cite{jiang2024fisherrf}. 

Let us consider EIG is given by $\mathcal{I}(\bx^{acq};\mathcal{X}) \in \R$, where $\bx^{acq}$ is the pose of the additional camera pose in SE(3) and $\mathcal{X}$ is the current collection of 3D Gaussians. The value is the expected information gain to the model if we introduce the view $\bx^{acq}$ and its ground truth $\by^{acq}$ into our training set. The NBV selection problem can be formulated as the following optimization objective:
\begin{equation}\label{eq:opt_ori} 
\underset{\bx^{acq} \in D^{pool}}{\arg\max} \mathcal{I}(\bx^{acq}; \mathcal{X}), \end{equation}
where $D^{pool}$ is the set of candidate camera poses, randomly generated around the waypoints. A detailed explanation of how $\mathcal{I}(\bx^{acq}; \mathcal{X})$ is computed can be found in the Appendix.

However, selecting the most informative view based on $\mathcal{X}$ is not necessarily optimal for enhancing the risk assessment of the payload robot's navigation. Therefore, we aim to select the next best view using the RaEM environment instead, such that the optimization becomes:
\begin{equation} \label{eq:opt_raem} 
\underset{\bx^{acq} \in D^{pool}}{\arg\max} \mathcal{I} \left(\bx^{acq}; \mathcal{M}({\bf p}) \right), 
\end{equation}
where $\mathcal{M}({\bf p})$ denotes the 3D Gaussians preserved in the masked environment.

%%%%%%%%%%%%%%%%%%%%%%%%%%%%%%%%%%%%%%%%%%%%%%%%%%%%%%%%%%%%%%%%%%%%%%%%%%%%%%%%%%%%%
\section{Experiments}   \label{sec:experiments}
As shown in Fig. \ref{fig:diag}, our experiments involve a payload robot positioned on the ground within a simulated environment, along with an aerial camera robot capable of maneuvering, capturing novel views, and sharing them with the payload robot.

\begin{figure*}[t]
    \centering
    \includegraphics[width=\linewidth]{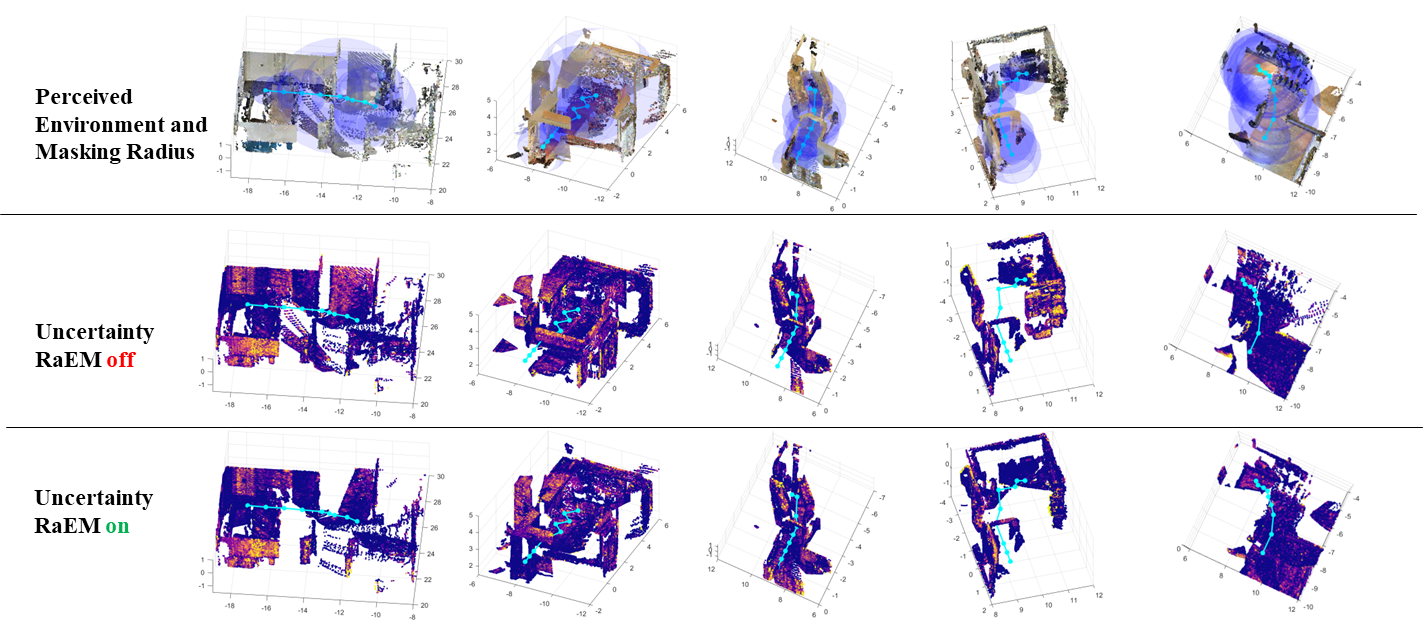}
    \caption{Perceived map and uncertainty map among various scenes of Matterport3D. The predefined path is shown in aqua color, and a darker color in uncertainty map \cite{jiang2024fisherrf} indicates less uncertainty. The result implies that RaEM helps the NBV algorithm to focus only on the safety-critical parts of the environment to reduce the reconstruction uncertainty and reveal more environments that are safety critical.}
    \label{fig:env_compare_i}
\end{figure*}

\begin{figure*}[t]
    \centering
    \includegraphics[width=0.8\linewidth]{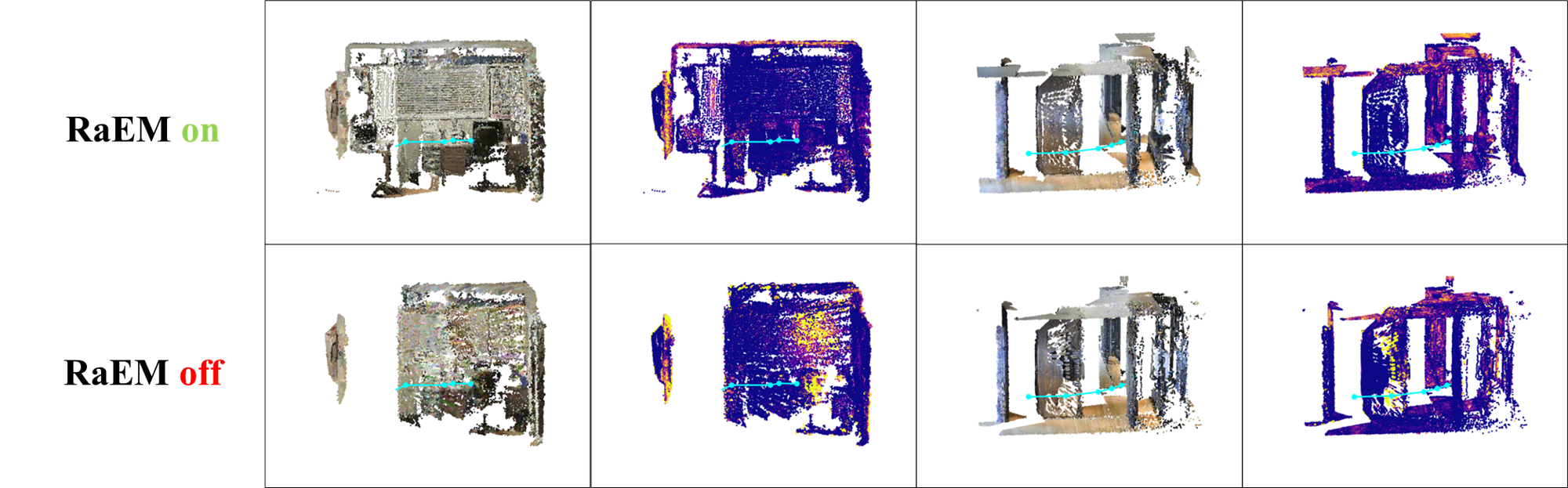}
    \caption{Uncertainty map comparison in the safety-critical region.}
    \label{fig:hl_compare}
\end{figure*}

\subsection{Experiment Setup}
The simulated environment is created using the Habitat simulator \cite{habitat19iccv}, with scenes from the Matterport3D \cite{Matterport3D} dataset for our experiments. We utilize the extensions of SplaTAM \cite{keetha2024splatam} for 3D Gaussian Splatting \cite{kerbl3Dgaussians} to construct the scene representation when using FisherRF \cite{jiang2024fisherrf}, and implicit occupancy network when using Ray Entropy \cite{yan2023activeimplicit} for NBV selection. In each scenario, the payload robot follows a predefined path, $\bp$, consisting of 10 waypoints to deliver the asset. Given the constraints on task completion time and computational resources, the camera robot is allowed to capture a novel view only upon reaching a candidate viewpoint. In total, the camera robot is permitted to capture 11 views to train and update the model. The candidate views are generated in three stages:
\begin{itemize} 
    \item \underline{Stage 1}: The payload robot is initially provided with a fixed view. Select 5 views, with a total of 250 candidate views randomly generated within a radius of 2 along the path $\bp$. 
    \item \underline{Stage 2}: Select 5 views, with a total of 250 candidate views randomly generated within a radius of 2 relative to the center of the explored environment. 
\end{itemize}

The payload robot uses the first 1+5 views to construct the initial environment representation during Stages 1. Then, it uses the next 5 views to refine the current model in Stage 2. At each stage, the NBV is selected by maximizing the expected information gain, as defined in \eqref{eq:opt_ori}. In Stage 2, when RaEM is activated, the NBV is chosen by solving \eqref{eq:opt_raem} within the masked environment.

\subsection{Implementation Details}

The camera robot captures an RGB-D observation of the environment with a resolution of $256 \times 256$, and the onboard model with 3D Gaussian Splatting is trained for $60$ iterations with the initial view. The view selection algorithm is applied to the randomly generated candidate views, where the Fisher Information on 3D Gaussian Splatting \cite{kerbl3Dgaussians} is computed using customized CUDA kernels, as in \cite{jiang2024fisherrf}. The log-prior regularizer $\lambda$ in \eqref{eq:hessian-reg} is set to $0.1$, and the depth loss weight $\gamma$ is set to $0.5$ across all experiments. Once the next view is selected, the camera robot uses the pathfinder provided by the Habitat API to determine the shortest action sequence to the target viewpoint. The onboard model is then optimized for $60$ iterations with each collected RGB-D observation during Stages 2 and 3, resulting in a total of $660$ iterations per experiment. When using RaEM, $\AVAR$ is evaluated with a confidence level of $\varepsilon = 0.1$, and the masking parameters are set to $\beta_1 = 0.2$ and $\beta_2 = 1$ for all experiments.

\subsection{Effectiveness of RaEM}

\renewcommand{\arraystretch}{1.2}
\begin{table}[t]
    \centering
    \begin{tabularx}{\linewidth}{c|X|X|X|X} 
        \toprule
        \multicolumn{1}{c|}{\textbf{NBV Policy}} 
        & \multicolumn{2}{c|}{\text{FisherRF} \cite{jiang2024fisherrf}} 
        & \multicolumn{2}{c}{\text{Ray Entropy} \cite{yan2023activeimplicit}} \\ 
        \cmidrule(lr){1-1} \cmidrule(lr){2-3} \cmidrule(lr){4-5}
        \textbf{Scene ID} & \text{Original} & \textbf{Ours} 
        & \text{Original} & \textbf{Ours} \\ 
        \midrule
        YVUC4YcDtcY & 0.8021 & \textbf{0.5770} & 0.3828 & \textbf{0.3413} \\ 
        2t7WUuJeko7 & 0.4667 & \textbf{0.4063} & 0.4663 & \textbf{0.3693} \\ 
        ARNzJeq3xxb & 0.5591 & \textbf{0.5113} & 0.4501 & \textbf{0.4119} \\
        RPmz2sHmrrY & 0.5266 & \textbf{0.2335} & 0.5791 & \textbf{0.3024} \\
        Vt2qJdWjCF2 & 0.6960 & \textbf{0.6565} & 0.6418 & \textbf{0.6145} \\
        q9vSo1VnCiC & 0.9547 & \textbf{0.6706} & 0.9235 & \textbf{0.7430} \\
        WYY7iVyf5p8 & 0.4547 & \textbf{0.4399} & 0.2111 & \textbf{0.1682} \\
        pa4otMbVnkk & 0.7471 & \textbf{0.5644} & 0.3959 & \textbf{0.3712} \\
        fzynW3qQPVF & 1.0088 & \textbf{0.8004} & 1.0106 & \textbf{0.8408} \\
        yqstnuAEVhm & 0.6889 & \textbf{0.3698} & 0.5730 & \textbf{0.4785} \\
        \midrule
        \textbf{$\mathbb{W}_2$ reduction}
        & \multicolumn{2}{c|}{$24.26\%$} & \multicolumn{2}{c}{$17.63\%$} \\
        \bottomrule
    \end{tabularx}
    \caption{Comparison of $\mathbb{W}_2(\Pro, \hat{\Pro}) \downarrow$ and $\mathbb{W}_2(\Pro^{\textup{ RaEM}}, \hat{\Pro}) \downarrow$ across various NBV policies and scenes from Matterport3D.}
    \label{tab:risk_dist_raem}
\end{table}

By shifting the focus of the NBV selection to the masked environment with RaEM, the model prioritizes utilizing the limited novel views on the safety-critical subset of the environment, as shown in Fig. \ref{fig:env_compare_i}. To demonstrate the effectiveness of RaEM, we evaluate the distribution of the closest distance\footnote{The closest point is determined in terms of $\scalebox{0.8}{$\AVAR$}_\varepsilon$, which can be quantified as $\displaystyle \dd \big(p_k,~ \argmin_{x_i} \scalebox{0.8}{$\AVAR$}_{\varepsilon}(\dd(p, x_i)) \, \big)$.} from the payload robot to the environment. The resulting distributions are denoted as $\Pro^{\textup{RaEM}}$ and $\Pro$ for the scenarios with and without RaEM, respectively. Additionally, we quantify the distribution of the closest distance to the environment using the ground truth point clouds, represented as $\hat{\Pro}$, which is directly obtained from the Matterport3D dataset, denoted by $\hat{x}_i \in \hat{\mathcal X}$. It is assumed that $\hat{x}_i$ follows a normal distribution
$
    \hat{x}_i \sim \N \left(\hat{\mu}_i,  \hat{\sigma}_i^2\right),
$
where $\hat{\mu}_i \in \R^3$ represents the ground truth position, and $\hat{\sigma}_i$ is a positive real value close to zero.

The Wasserstein distance measures the distance between two probability distributions, and the type-2 Wasserstein distance \cite{givens1984class} between two normal distributions $\Pro$ with parameters $(\mu, \sigma)$ and $\hat{\Pro}$ with parameters $(\hat{\mu}, \hat{\sigma})$ is given by
\[
    \mathbb{W}_2(\Pro, \hat{\Pro}) = \sqrt{\|\mu - \hat{\mu}\|_2^2 + (\sigma - \hat{\sigma})^2},
\]
where $\Pro$ represents the distribution of the closest distance to the perceived point clouds (with or without RaEM), and $\hat{\Pro}$ is the ground truth distribution of the closest distance. This quantity measures the deviation of the reconstructed 3D scene representation from the ground truth environment, particularly in the safety-critical regions. The effectiveness of RaEM is demonstrated in Table \ref{tab:risk_dist_raem}, where we compare $\mathbb{W}_2(\Pro_i^{\textup{ RaEM}}, \hat{\Pro}_i)$ and $\mathbb{W}_2(\Pro_i, \hat{\Pro}_i)$ for both FisherRF and Ray Entropy, indicating that the proposed RaEM framework is also algorithm-agnostic.

\begin{figure}[t]
    \centering
    \includegraphics[width=\linewidth]{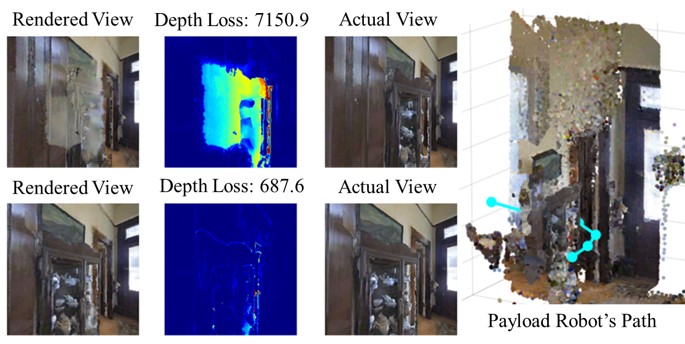}
    \caption{View and depth comparison in the safety-critical region. Top: Views and depth with RaEM on. Bottom: Views and depth without RaEM.}
    \label{fig:view_compare}
\end{figure}

The effectiveness of RaEM is further demonstrated in Fig. \ref{fig:hl_compare}, in this case, RaEM navigates the NBV algorithm to select the views that reduces the reconstruction uncertainty of the safety-critical region, i.e., the wall near the predefined path. On the other hand, the non risk-aware NBV algorithm selects the views solely based on the entire perceived environment, which results in a significantly higher uncertainty in the safety-critical region.  The similar effect is observed in Fig. \ref{fig:view_compare}, where, in the safety-critical regions, the NBV algorithm with RaEM achieves better image rendering quality and more accurate depth estimation.

\subsection{Dynamic Masking via Risk-awareness}

\renewcommand{\arraystretch}{1.2}
\begin{table}[t]
    \centering
    \begin{tabular}{ c|c|c|c|c|c|c } 
        \toprule
        \textbf{Method} & \multicolumn{5}{c|}{\textbf{Uniform Radius}} & \textbf{RaEM} \\ 
        \midrule
        $r_{\text{mask}}(k)$ & $3$ & $2$ & $1$ & $0.5$ & $0.1$ & $\beta_1 e^{-\beta_2 \alpha_k}$ \\
        \midrule
        $\mathbb{W}_2(\Pro, \hat{\Pro})  \downarrow$ & 0.488 & 0.494 & 0.559 & 0.392 & N/A & \textbf{0.370} \\
        \bottomrule
    \end{tabular}
    \caption{$\mathbb{W}_2$ comparison of masking radius between RaEM (dynamic) and uniform radius (static).}
    \label{tab:risk_dist_radius}
\end{table}

\begin{figure}[t]
    \centering
    \includegraphics[width=\linewidth]{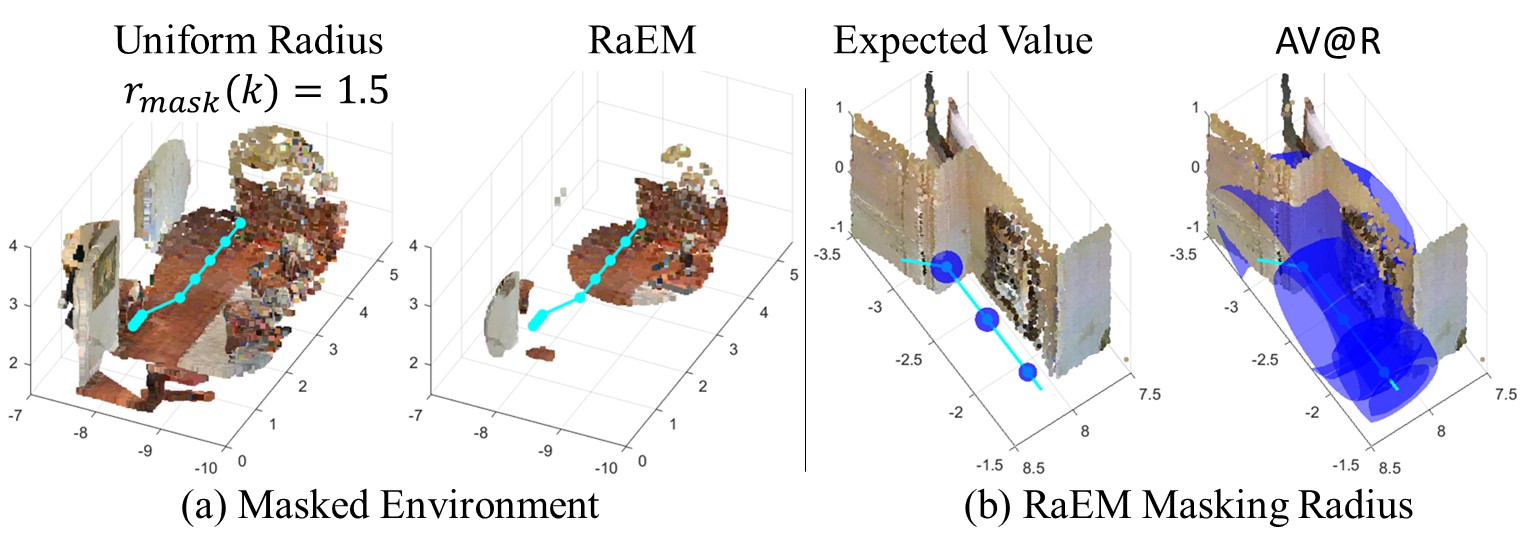}
    \caption{Comparison of masked environment and masking radius. }
    \label{fig:radius_compare}
\end{figure}

In addition to emphasizing the safety-critical proximity of the payload robot, RaEM offers dynamic adjustments of the masking radius based on the risk level of the current waypoint. A straightforward way to highlight the safety-critical region is by using a constant uniform masking radius along the path $\bp$. For instance, the environment can be masked by setting $r_{\text{mask}}(k) = 1.5$ in \eqref{eq:masking}. As shown in Fig. \ref{fig:radius_compare} (a), RaEM effectively concentrates more on the safety-critical regions, whereas a uniform radius treats all 3D Gaussians equally within its proximity. This observation is further validated in Table \ref{tab:risk_dist_radius} with different masking radii.

Other common risk measures such as the expected value $\mathbb{E} [\dd(p_k, x_i)]$ can also be used for dynamic environment masking via \eqref{eq:radius}. However, when the predefined path is close to the uncertain environment, $\mathbb{E} [\dd(p_k, x_i)]$ tends to underestimate the collision risk, as it fails to account for the safety-critical regions. This is illustrated in Fig. \ref{fig:radius_compare} (b).

%%%%%%%%%%%%%%%%%%%%%%%%%%%%%%%%%%%%%%%%%%%%%%%%%%%%%%%%%%%%%%%%%%%%%%%%%%%%%%%%%%%%%
\section{Conclusion}

In this paper, we introduce a risk-aware active view acquisition framework aimed at improving risk and safety assessment for robot navigation while advancing 3D scene reconstruction. Our proposed approach extends the uncertainty quantification in FisherRF to risk measures, allowing RaEM to strategically navigate the NBV algorithm to the safety-critical regions of the unknown environment. By maximizing the expected information gain in the dynamically characterized safety-critical region, the RaEM framework ensures more accurate risk assessment for robot navigation while concentrating the limited sensing capability on the task-related region of the environment. Extensive experimental results validate the effectiveness of our method in near real-world scenarios, showcasing its potential to significantly advance robotic systems and autonomous navigation. This work marks a substantial step toward developing a more resilient and safety-oriented framework for active view acquisition and comprehensive 3D scene reconstruction.

%%%%%%%%%%%%%%%%%%%%%%%%%%%%%%%%%%%%%%%%%%%%%%%%%%%%%%%%%%%%%%%%%%%%%%%%%%%%%%%%%%%%%
\section*{Appendix}
\subsection{Risk Quantification with Anisotropic Covariance Matrix}

In the case when the covariance matrix $\Sigma_i$ of the 3D Gaussian $x_i$ is anisotropic, the distance between a given waypoint and each 3D Gaussian point can be computed as follows.

\begin{lemma}   \label{lem:prob_dist_d(x,p)_aniso}
The signed distance, $\dd(p, x_i)$, between a fixed waypoint, $p \in \R^3$, and any 3D Gaussian point, $x_i \sim \N(\mu_i,\Sigma_i)$, follows a normal distribution such that
\[
    \dd(p,x_i) \sim \N \left(\|\mu_i - p\|_2, \, \left(\frac{\mu_i - p}{\|\mu_i - p\|_2} \right)^T \Sigma_i \, \frac{\mu_i - p}{\|\mu_i - p\|_2}\right).
\]
\end{lemma}
Then, the quantification of $\AVAR(\dd(p,x_i))$ follows by substituting $\sigma^2$ with $\left(\frac{\mu_i - p}{\|\mu_i - p\|_2} \right)^T \Sigma_i \, \frac{\mu_i - p}{\|\mu_i - p\|_2}$ in \eqref{eq:avar}.

\subsection{Alternative Risk Evaluations}

While $\AVAR_\varepsilon$ effectively captures the collision risk for the payload robot, a more intuitive representation of this risk can be created by assigning a value within $[0,\infty]$ to quantify the ``dangerousness'' of the scenario. This approach uses a family of level sets, defined by 
\begin{equation} 
C_{\delta} = \Big(-\infty, ~ \frac{d_s}{1+\delta}\Big),
\end{equation}
where $d_s > 0$ is a predefined cutoff distance. Here, the payload robot is deemed safe if $\dd(p,x_i) \geq d_s$ or $\dd(p,x_i) \notin C_0$, and a collision is defined as when $\dd(p,x_i) < 0$ or $\dd(p,x_i) \in C_{\infty}$. Accordingly, the associated collision risk, as defined in \cite{liu2023risk}, is
\begin{align*}  
    \mathcal{A}_{\varepsilon}({\textup{d}}(p,x_i)) = \sup \left\{ \delta \geq 0 \,\Big|\, \AVAR_\varepsilon({\textup{d}}(p,x_i)) \in C_\delta \right\}.
\end{align*}

\begin{corollary}  \label{cor:risk_levelset}
     For any given waypoint $p$ and a 3D Gaussian $x_i \sim \N(\mu_i,\sigi^2 I_3)$, the risk of collision with respect to the level sets is 
    \[
        \mathcal{A}_{\varepsilon}({\textup{d}}(p,x_i)) :=\begin{cases}
            0, &\text{if} ~ \frac{\mu - d_s}{\sigi} \geq \kappa_{\varepsilon}\\
            \dfrac{d_s}{\mu-\kappa_\varepsilon \sigi} - 1, &\text{if} ~ \kappa_{\varepsilon} \in \big(\frac{\mu-d_s}{ \sigi}, \frac{\mu}{\sigi}  \big) \\
            \infty, &\text{if} ~ \frac{\mu}{\sigi} \leq  \kappa_{\varepsilon}
            \end{cases},
    \]
    where $\mu = \|\mu_i - p\|_2$, $\kappa_{\varepsilon} = \left(\sqrt{2\pi} \varepsilon \exp(\io^2) \right)^{-1}$, and  $\io = \textup{erf}^{-1} (2\varepsilon-1)$.  
\end{corollary}

This new risk representation, as described above, quantifies the danger similarly to Theorem~\ref{thm:avar} but offers a more intuitive understanding. A higher value of $\mathcal{A}_{\varepsilon}$ directly translates to a greater chance of a collision and potentially more severe consequences. The case $\mathcal{A}_{\varepsilon}(y)=0$ indicates that $\AVAR_\varepsilon(y)$ is on the boundary of, or outside, $C_0$, and the robot is considered safe. We observe that $\mathcal{A}_{\varepsilon}(y) > 0$ if and only if $\AVAR_\varepsilon(y) \in C_{\delta}$ for some $\delta >0$. In the extreme scenario where $\mathcal{A}_{\varepsilon}(y)=\infty$, it signifies that $\AVAR_\varepsilon(y)$ is located within $C_{\infty}$.

\subsection{Quantification of the Fisher Information Gain}
\begin{figure}
    \centering
    \includegraphics[width=\linewidth]{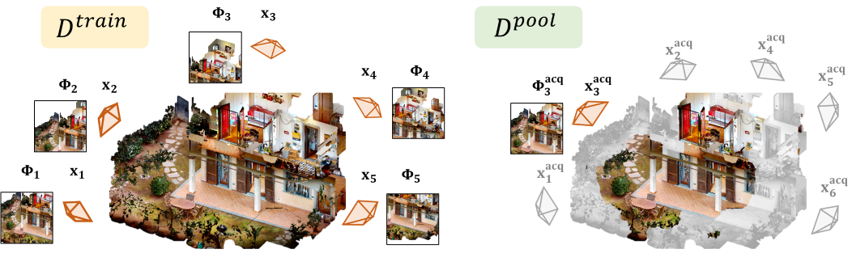}
     \caption{Observation pairs $(\bx \in D^{train}, \by)$ and candidate observation pairs $(\bx^{acq} \in D^{pool}, \by^{acq})$.}
    \label{fig:train_vs_acq}
\end{figure}

The Fisher Information serves as a measure of the information that an observation pair $(\bx, \by)$ carries about the unknown parameters $\bw$ that define the model $p(\by|\bx; \bw)$. In the context of novel view synthesis, the pair $(\bx, \by)$ correspond to the camera pose $\bx$ and the image observation $\by$ at that pose, respectively. Here, $\bw$ represents the parameters of the radiance field. The primary objective in neural rendering is to minimize the negative log-likelihood (NLL) between the rendered images and ground truth images in the holdout set. This minimization is inherently indicative of the scene reconstruction quality, as defined by
\begin{equation}    \label{eq:logp}
   - \log \Pro(\by|\bx, \bw) = \left(\by - f(\bx, \bw)\right)^T \left(\by - f(\bx, \bw)\right),
\end{equation}
where $f(\bx, \bw)$ represents the onboard rendering model with 3D Gaussians. Under the regularity conditions \cite{schervish2012theory}, the Fisher Information corresponding to the model $\log \Pro(\by|\bx; \bw)$ is characterized as the Hessian of the log-likelihood function with respect to the model parameters $\bw$:
\begin{equation*}
\scalebox{0.95}{$
    \mathcal{I}(\bw) = - \mathbb{E}_{p(\by|\bx, \bw)} \left[\dfrac{\partial^2 \log p(\by|\bx, \bw)}{\partial \bw^2} \Bigg| \bw \right] = \bH''[\by|\bx, \bw],
$}
\end{equation*}
where $\bH''[\by|\bx, \bw]$ is the Hessian matrix of \eqref{eq:logp}.

In the context of the active view selection problem, we initiate the process with a training set $D^{train}$ and obtain an initial estimation of parameters $\bw^*$ using $D^{train}$. Our main objective is to identify the NBV that maximizes the Expected Information Gain~\cite{lindleyEIG, HoulsbyBAL,KirschBatchBALD} among potential viewpoints $\bxa \in D^{pool}$ in comparison to $D^{train}$, where $D^{pool}$ denotes the collection of available views:
\begin{align}
    & \mathcal{I}[\bw^*; \{\bya\}|\{\bxa\}, D^{train}] \nonumber \\
    & = H[\bw^* | D^{train}] - H [\bw^*| \{\bya\}, \{\bxa\}, D^{train}] 
    \label{eq:entropy-acq}
\end{align}
where $H[\cdot]$ is the entropy \cite{kirsch2022unifying}. The relation between $D^{train}$ and $D^{pool}$ is also depicted in Fig. \ref{fig:train_vs_acq}.

Given the log-likelihood's expression as in \eqref{eq:logp}, specifically the rendering error in our context, the entropy difference on the right-hand side of \eqref{eq:entropy-acq} can be estimated as follows \cite{kirsch2022unifying}: 
% \begin{equation}    \label{eq:entropy-log}
%     \scalebox{0.95}{$
%     \begin{aligned}
%         \frac{1}{2} \log \det \left(\bH''[\{\bya\}|\{\bxa\}, \bw^*]\; \bH''[\bw^*| D^{train}]^{-1} + I \right) \nonumber\\
%         \leq \frac{1}{2}  \tr\left( \bH''[\{\bya\}|\{\bxa\}, \bw^*]\; \bH''[\bw^*| D^{train}]^{-1}\right).
%     \end{aligned}
%     $}
% \end{equation}
\begin{align*}   \label{eq:entropy-log}
 \frac{1}{2} \log \det \left(\bH''[\{\bya\}|\{\bxa\}, \bw^*]\; \bH''[\bw^*| D^{train}]^{-1} + I \right) \nonumber\\
    \leq \frac{1}{2}  \tr\left( \bH''[\{\bya\}|\{\bxa\}, \bw^*]\; \bH''[\bw^*| D^{train}]^{-1}\right).
\end{align*}
Since Fisher Information is additive, $\bH''[\bw^*| D^{train}]^{-1}$ can be computed by aggregating the Hessian of model parameters from all viewpoints in $\{D^{train}\}$ prior to inversion. The forthcoming optimal view $\bxa$ is selected by:
\begin{equation}    \label{eq:opt}
    \argmax_{\bxa} \tr \Big( \bH''[\bya | \bxa, \bw^*] \; \bH''[\bw^*| D^{train}]^{-1} \Big).
\end{equation}
The model's Hessian $\bH''[\by|\bx, \bw^*]$ is determined as:
\begin{equation}    \label{eq:hessian-full}
    \begin{aligned}
        &\bH''[\by|\bx, \bw^*] =\\
        & \hspace{2mm} \nabla_\bw f(\bx; \bw^*) ^T \nabla_{f(\bx;\bw^*)}^2 H[\by| f(\bx;\bw^*)] \nabla_\bw f(\bx; \bw^*)
    \end{aligned}
\end{equation}
where $\bH''[\by| \bx,\bw^*]$ in our case is equal to the covariance of the RGB measurement that we set equal to one.
Thus, the Hessian matrix can be computed solely using the Jacobian matrix of $f(\bx, \bw)$:
\begin{equation}\label{eq:hessian-simplified}
    \bH''[\by|\bx, \bw^*] = \nabla_\bw f(\bx; \bw^*) ^T \nabla_\bw f(\bx; \bw^*).
\end{equation}

This optimization of the objective in \eqref{eq:opt} is feasible without the ground truth of the candidate views' ground truths $\{\bya\}$, as expected since Fisher Information is independent of the observations.
The Hessian in \eqref{eq:hessian-simplified} possesses a limited number of non-diagonal elements since each pixel's independence is considered in $- \log p(\by|\bx, \bw)$.
Moreover, recent NeRF models \cite{fridovich2022plenoxels,muller2022instant, reiser2021kilonerf,sun2022direct} typically use structured local parameters whereby each parameter influences only the radiance and density within a confined spatial domain, aiding in expedited convergence and rendering. Thus, only parameters affecting the pixel colors have non-zero entries in the Hessian matrix $ \bH''[\by|\bx, \bw^*]$. 
Nevertheless, with the parameter count often exceeding 200 million, computing without sparsification or approximation is unfeasible. Practically, the Laplace approximation method \cite{laplace2021, bayesian-interpolation} is employed, approximating the Hessian matrix with its diagonal components and a log-prior regularizer $\lambda I$:
\begin{equation}    \label{eq:hessian-reg}
    \scalebox{0.95}{$
        \bH''[\by|\bx, \bw^*] \approx \diag(\nabla_\bw f(\bx, \bw^*) ^T \nabla_\bw f(\bx, \bw^*)) + \lambda I.
    $}
\end{equation}
This approximation further simplifies the inversion of the Hessian matrix $ \bH ''[\by|\bx, \bw^*]$.

\subsection{Proofs of Theoretical Results}

\noindent \underline{\textbf{Proof of Lemma \ref{lem:prob_dist_d(x,p)}:}} The result is immediate by evaluating the marginal distribution of $(x_i - p) \sim \N(\mu_i - p, \sigi I_3)$ along the vector $\frac{\mu_i - p}{\|\mu_i - p\|_2}$ such that the marginal mean is 
\begin{equation*}
        \big(\frac{\mu_i - p}{\|\mu_i - p\|_2} \big)^T (x_i - p) = \|\mu_i - p\|_2,
\end{equation*}
and the variance 
\begin{equation*}
        \hspace{2cm} \big(\frac{\mu_i - p}{\|\mu_i - p\|_2} \big)^T \sigi^2 I_3 \, \frac{\mu_i - p}{\|\mu_i - p\|_2} = \sigi^2. \hspace{2cm} \square
\end{equation*}

\vspace{1mm}

\noindent \underline{\bf Proof of Theorem \ref{thm:avar}:} Given the fact that the position of the 3D Gaussian point $x_i$ follows a normal distribution, one has
\begin{equation*}
    \scalebox{0.95}{$
        \Pro \{\dd(p,x_i) < z\} = \frac{1}{2} \left(1+\textup{erf} \big(\frac{z-\tilde{\mu}}{\sqrt{2} \sigi} \big) \right),
    $}
\end{equation*}
and considering the fact that the probability density function of the normal distribution is continuous,
\[
    \VAR_\varepsilon (\dd(p,x_i)) = \sqrt{2} \sigi \, \textup{erf}^{-1}(2\varepsilon-1) + \|\mu_i - p\|_2.
\]
The corresponding $\AVAR$ is evaluated by using change of variable such that
\begin{equation*}
    \scalebox{0.85}{$
    \begin{aligned}
        \AVAR_\varepsilon (\dd(p,x_i)) &= \frac{1}{\varepsilon} \int_{-\infty}^{\VAR_\varepsilon (\dd(p,x_i))} z \, \frac{1}{\sqrt{2 \pi} \, \sigi} \exp \left( -\frac{(z-\mu)^2}{2 \sigi^2}\right)\textup{d}z\\
        & = \mu - \frac{\sigi}{\sqrt{2\pi} \, \varepsilon \exp(\io^2)},
    \end{aligned}
    $}
\end{equation*}
where $\mu = \|\mu_i - p\|_2$ and $\io = \textrm{erf}^{-1} (2\varepsilon-1)$.
\hfill$\square$

\vspace{1mm}
    
\noindent \underline{\bf Proof of Corollary \ref{cor:risk_levelset}:} Given that the probability density function of $\dd(p,x_i)$ is continuous, one can represent the risk in terms of level sets by considering when $\AVAR_\varepsilon (\dd(p,x_i)) > \frac{r}{c}$, $\AVAR_\varepsilon (\dd(p,x_i)) < 0$, and $\AVAR_\varepsilon(\dd(p,x_i)) \in (0,\frac{r}{c})$. Then, middle branch is obtained by solving for $\AVAR_\varepsilon(\dd(p,x_i)) = \frac{r}{\delta+c}$.
\hfill$\square$

\vspace{1mm}

\noindent \underline{\textbf{Proof of Lemma \ref{lem:prob_dist_d(x,p)_aniso}:}} The proof of Lemma \ref{lem:prob_dist_d(x,p)_aniso} follows the similar lines of argument as in the proof of Lemma \ref{lem:prob_dist_d(x,p)} by replacing the isotropic $\sigma^2 I_3$ with the anisotropic covariance matrix $\Sigma_i$ in the computation of the marginal variance.
\hfill$\square$

\printbibliography

\end{document}